\documentclass[11pt,a4paper]{article}
\usepackage[hyperref]{acl2022}

\usepackage{etoolbox}
\newtoggle{release}
\newtoggle{4page}
\toggletrue{release}

\iftoggle{release}{}{
    \toggletrue{4page}  
}

\hypersetup{
 	pdfinfo={
 		Title={MAUD: An Expert-Annotated Legal NLP Dataset for Merger Agreement Understanding},
 		Author={Steven H. Wang and Antoine Scardigli and Leonard Tang and Dimitry Levkin and Anya Chen Spencer Ball and and Wei Chen and Thomas Woodside and Oliver Zhang and Dan Hendrycks},
 		Subject={Legal NLP},
 		Keywords={merger agreements, contract review, bert}
 	}
}

\usepackage{times}
\usepackage{latexsym}

\usepackage{graphicx}
\usepackage{enumitem}
\usepackage{microtype}
\usepackage{subcaption}
\usepackage{textcomp}
\usepackage{booktabs}
\usepackage{gensymb}
\usepackage{mathtools}
\usepackage{amssymb}
\usepackage{wrapfig}
\usepackage{stfloats}
\usepackage{multirow}
\usepackage{tabularx}
\usepackage{todonotes}
\usepackage{needspace}
\newcolumntype{Y}{>{\centering\arraybackslash}X}
\newcolumntype{L}{>{\raggedright\arraybackslash}X}

\usepackage{multirow}
\usepackage{arydshln}
\usepackage{makecell}
\usepackage{cleveref}
\usepackage{wasysym}

\newif\ifcomments
\ifcomments
    \providecommand{\eric}[1]{{\protect\color{magenta}{[Eric: #1]}}}
\else
    \providecommand{\eric}[1]{}
\fi

\title{MAUD: An Expert-Annotated Legal NLP Dataset for\\Merger Agreement Understanding}

\author{\makecell{Steven H. Wang$^1$\thanks{~~Correspondence to  \href{mailto:stewang@student.ethz.edu}{\tt stewang@student.ethz.edu}} \hspace{0.3cm} Antoine Scardigli$^{1}$ \hspace{0.3cm} Leonard Tang$^2$ \hspace{0.3cm} Wei Chen$^3$ \hspace{0.3cm} Dimitry Levkin$^3$
\\
\hspace{0.3cm} Anya Chen$^4$  \hspace{0.3cm} Spencer Ball$^5$ \hspace{0.3cm} Thomas Woodside$^6$ \hspace{0.3cm} Oliver Zhang$^7$ \hspace{0.3cm} Dan Hendrycks$^8$} \\
$^1$ETH Zürich  $^2$Harvard University $^3$The Atticus Project $^4$The Nueva School \\
$^5$University of Wisconsin, Madison $^6$Yale University $^7$Stanford University $^8$UC Berkeley
\\ 
}

\begin{document}
\maketitle
\begin{abstract}
Reading comprehension of legal text can be a particularly challenging task due to the length and complexity of legal clauses and a shortage of expert-annotated datasets. To address this challenge, we introduce the Merger Agreement Understanding Dataset (MAUD), an expert-annotated reading comprehension dataset based on the American Bar Association's 2021 Public Target Deal Points Study,
with over 39,000 examples and over 47,000 total annotations. 
Our fine-tuned Transformer baselines show promising results, with models performing well above random on most questions. However, on a large subset of questions, there is still room for significant improvement.
As the only expert-annotated merger agreement dataset, MAUD is valuable as a benchmark for both the legal profession and the NLP community.
\end{abstract}

\section{Introduction}

While pretrained Transformers \cite{Devlin2019BERTPO, Brown2020LanguageMA} have surpassed humans on reading comprehension tasks such as SQuAD 2.0 \cite{Rajpurkar2018KnowWY} and SuperGLUE \cite{Wang2019SuperGLUEAS}, their accuracy in understanding real-world specialized legal texts remains underexplored.

Reading comprehension of legal text can be a particularly challenging natural language processing (NLP) task due to the length and complexity of legal clauses and the difficulty of collecting expert-annotated datasets.
To help address this challenge, we introduce the Merger Agreement Understanding Dataset (MAUD),
a legal reading comprehension dataset curated under the supervision of highly specialized mergers-and-acquisitions (M\&A) lawyers and used in the American Bar Association's 2021 Public Target Deal Points Study (``ABA Study").
\iftoggle{release} {
The dataset and code for MAUD
can be found at \href{http://github.com/TheAtticusProject/maud}{github.com/TheAtticusProject/maud}.
} {
The dataset and baseline training code for MAUD
and the original merger agreement texts will be published to GitHub.
}


Public target company acquisitions are the most prominent business transactions, valued at hundreds of billions of dollars each year. Merger agreements are the legal documents that enable these acquisitions, and key clauses in these merger agreements are called ``deal points."


Lawyers working on the ABA Study perform contract review on merger agreements. In general, contract review is a two-step process. First, lawyers extract key legal clauses from the contract (a span extraction task). Second, they interpret the meaning of these legal clauses (a reading comprehension task). In the ABA Study, the lawyers extract deal points from merger agreements, and for each deal point 
they answer a set of standardized multiple-choice questions.


Models trained on MAUD's expert-annotated data can learn to answer 92 reading comprehension questions from the 2021 ABA Study, given extracted deal point text from merger agreements.
By answering these questions, models interpret the meaning of specialized legal language and categorize the different agreements being made by companies in the contract.

Span extraction and reading comprehension are both important and challenging tasks in legal contract review.
A large-scale expert-annotated span extraction benchmark for contract review is already available in \citet{hendrycks_cuad_2021}. However, to the best of our knowledge, there is no large-scale expert-annotated reading comprehension dataset for contract review or any other legal task in the English language. Therefore in this short paper, we focus on the legal reading comprehension task. (Appendix \ref{appendix:extraction} presents a preliminary benchmark for the extraction task for interested researchers.)


Annotating MAUD was a collective effort of over 10,000 hours by law students and experienced lawyers.
Prior to labeling, each law student attended 70-100 hours of training, including lectures and workshops from experienced M\&A lawyers.
Each annotation was labelled by three law student annotators, and these labels were verified by an experienced lawyer.
See Appendix \ref{appendix:labeling process} for more information on the annotation process.
We estimate the pecuniary value of MAUD to be over \$5 million using a prevailing rate of \$500 per hour in M\&A legal fees.



\begin{figure*}
    \centering
    \iftoggle{4page}{
        \includegraphics[width=16cm]{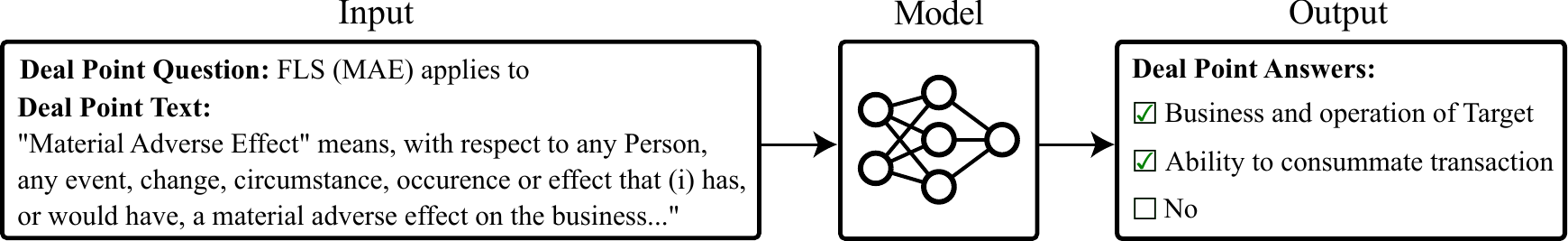}
    }{
       \includegraphics[width=16cm]{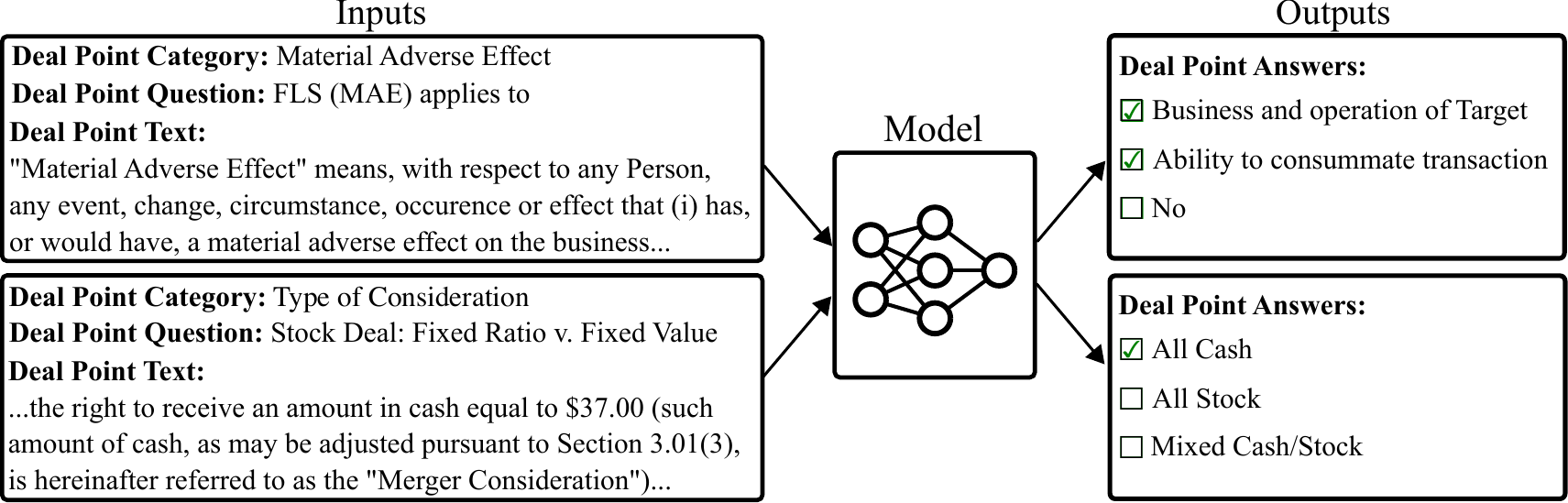} 
    }
    \caption{
    MAUD contains 39,000+ examples for 92 different reading comprehension questions about merger agreements.
    Given a \textit{deal point question} and \textit{deal point text}, a model learns to predict the correct answer(s) from a list of possible answers standardized by the 2021 ABA Study. The deal point texts above are truncated for display.}
    \label{fig:main}
\end{figure*}

\section{Related Work}
Due to the high costs of contract review and the specialized skills it requires, understanding legal text has proven to be a ripe area for NLP research.

\paragraph{Information Extraction for Legal NLP.}
One area of contract review research focuses on information extraction and document segmentation. \citet{Chalkidis2017ExtractingCE} introduce a dataset for extracting basic information from contracts, with follow-up modeling work using RNNs \citep{Chalkidis2018ObligationAP} and Transformers \citep{chalkidis_legal_2020}.
\citet{lippi_claudette_2019} introduce a small 
expert-annotated dataset for identifying ``unfair" clauses in 50 online terms of services. \citet{tuggener_ledgar_2020} introduce a semi-automatically constructed dataset of legal contracts for entity extraction. \citet{Leivaditi2020ABF} introduce an expert-annotated dataset of 2960 annotations for 179 lease agreements.
\citet{hendrycks_cuad_2021} introduce CUAD, an expert-annotated contract review dataset containing 13,010 annotations for 150 legal contracts.

 \paragraph{Reading Comprehension for Legal NLP.}
 \citet{koreeda_contract_2021} introduce a crowd-worker-annotated dataset containing 7191 Natural Language Inference questions about spans of non-disclosure agreements. \citet{Hendrycks2020MeasuringMM} propose a question-answering dataset sourced from freely available online materials, containing questions (including legal exam questions) from dozens of specialized areas.
\citet{zheng2021does_casehold} present a multiple-choice reading comprehension dataset with 53,317 annotations automatically extracted from US case law citations.
\citet{Duan2019CJRCAR} present a Chinese-language legal reading comprehension dataset, with about 50,000 expert-generated annotations of Chinese judicial rulings.
In our work we present a legal reading comprehension dataset with 47,457 expert-generated annotations about merger agreements. To the best of our knowledge, MAUD is the only English-language legal reading comprehension dataset that is both large-scale and expert-annotated.



\section{MAUD: A Legal NLP Dataset for Merger Agreement Understanding} \label{section:dataset}
MAUD consists of 47,457 annotations based on legal text extracted from 152 English-language public merger agreements.
MAUD's merger agreements were sourced from the EDGAR system maintained by the U.S. Securities and Exchange Commission.

\paragraph{Terminology.} \textit{Deal points} are legal clauses that define when and how the parties in a merger agreement are obligated to complete an acquisition.
We refer to the text of these clauses (extracted by annotators from merger agreements) as \textit{deal point text}s.
Multiple \textit{deal point questions} can be asked about each deal point text; the subset of applicable questions is determined by the text's \textit{type}. Each deal point question can be answered by one or more predefined \textit{deal point answers}. 


\begin{table*}[htbp]
\setlength\tabcolsep{11pt}
\centering
\scalebox{0.85}{
\begin{tabular}{l|ccc|c}
\textbf{Deal Point Category}  & \begin{tabular}{@{}c@{}}\textbf{Main} \\ \textbf{Dataset}\end{tabular} & \begin{tabular}{@{}c@{}}\textbf{Rare Answers} \\ \textbf{Dataset}\end{tabular}  & \begin{tabular}{@{}c@{}}\textbf{Abridged} \\ \textbf{Dataset}\end{tabular} & \begin{tabular}{@{}c@{}}\textbf{All} \\ \textbf{Datasets}\end{tabular} \\
\hline
Conditions to Closing & 3,411 & 298 & 4,052 & 7,761
 \\
Deal Protection and Related Provisions & 6,491 & 2,280 & 5,937 & 14,708
 \\
General Information & 152 & 17 & 173 & 342
 \\
Knowledge & 388 & 23 & 258 & 669
 \\
Material Adverse Effect & 8,816 & 871 & 3,273 & 12,960
 \\
Operating and Efforts Covenant & 1,216 & 191 & 1,054 & 2,461
 \\
Remedies & 149 & 0 & 181 & 330
 \\
 \hline
All Categories & 20,623 & 3,680 & 14,928 & 39,231
 \\
 \hline
\end{tabular}}

\caption{Number of MAUD examples contained in each dataset by category. Each example is a question-answer pair corresponding to an extracted deal point text.}
\label{table:dataset_size}
\end{table*}

\paragraph{Deal Points in MAUD.}
The deal points in MAUD are standardized by the 2021 ABA Study.
For the 2021 ABA Study, the American Bar Association appointed an M\&A attorney to design 130 deal point questions reflecting recent legal developments and deal trends of interest. Of the 130 different deal point questions in the 2021 ABA Study, 92 are represented in MAUD.

MAUD contains 8,226 unique deal point text annotations and 39,231 question-answer annotations (i.e. examples), for a total of 47,457 annotations. Each text belongs to one of 22 deal point types (see Table \ref{table:extraction_types}), and the deal point type determines the subset of questions that pertain to each text. The deal point types are further grouped into seven categories (see Appendix \ref{appendix:categories}) which we use for scoring purposes.


\paragraph{Task.} MAUD is a multiple-choice reading comprehension task. The model predicts the correct deal point answer from a predefined list of possible answers associated with each question. (See Figure \ref{fig:main} for an example). Several deal point questions in the ABA Study are in fact multilabel questions, but for uniformity we cast all multilabel questions as binary multiple-choice questions. This increases the effective number of questions from 92 to 144.


\subsection{MAUD Datasets and Splits}

MAUD contains three datasets (main, abridged, and rare answers) corresponding to three methods of generating examples.
See Table \ref{table:dataset_splits} for the number of examples contained in each dataset.

\paragraph{Main Dataset.} The main dataset contains 20,623 examples with original deal point text extracted from 152 merger agreements by expert annotators.

\paragraph{Abridged Dataset.} The abridged dataset contains 14,928 examples with deal point text extracted from 94 of the 152 merger agreements included in the main dataset. Texts in the abridged dataset are joined substrings of the main texts (with the delimiter ``\textless omitted\textgreater" indicating skipped text).
Because many texts contain answers to multiple questions, we provide the abridged data to guide a model to recognize the most pertinent text.

\paragraph{Rare Answers Dataset.} The rare answers dataset contains 3,680 examples that have rare answers to a question. Legal experts made small edits to texts in the main dataset to create deal points with rare answers. See Appendix \ref{appendix:labeling process} for an example edit.
We introduced the rare answers dataset to
ameliorate imbalanced answer distributions in the main dataset. In particular, some answers in the main dataset appear in fewer than 3 contracts, making a train-dev-test split impossible.

\begin{table}[htbp]
\centering
\begin{tabular}{|l|ccc|c|}
\hline
& \textbf{train} & \textbf{dev} & \textbf{test} & \textbf{overall} \\
\hline
\textbf{main} & 13,256 & 3,471 & 3,896 & 20,623 \\
\textbf{abridged} & 9,647 & 2,526 & 2,755 & 14,928 \\
\textbf{rare} & 2,924 & 756 & 0 & 3,680 \\
\hline
\textbf{overall} & 25,827 & 6,753 & 6,651 & 39,231 \\
\hline
\end{tabular}
\caption{The number of examples in MAUD, grouped by splits (train, dev, test) and by dataset (main, abridged, rare answers).}
\label{table:dataset_splits}
\end{table}

\paragraph{Train, Dev, and Test Splits.}
We construct the train-dev-test split as follows. We reserve a random 20\% of the combined main and abridged datasets as the test split. The remaining main and abridged examples are combined with the rare answers data, and then split 80\%-20\% to form the train and dev splits. All splits are stratified by deal point question-answer pairs.
Since each question belongs to one category, it follows that category proportions are also balanced across splits.

To avoid data leakage due to main dataset and abridged dataset examples having overlapping text and the same answer, we always split the main examples first and then place abridged examples from the same contract in the same split.

 \begin{table*}[htbp]
 \setlength\tabcolsep{11pt}
 \centering
 \scalebox{0.85}{\begin{tabular}{l|cccccc}
 \textbf{Deal Point Category} & \textbf{Random} & \textbf{BERT} & \textbf{RoBERTa} & \hspace{-0.1cm}\textbf{LegalBERT}\hspace{-0.1cm} & \textbf{DeBERTa} & \textbf{BigBird} \\
 \hline
 Conditions to Closing & 20.4\% & 41.7\% & 41.6\% & 32.0\% & \textbf{48.2\%} & 46.6\% \\
 Deal Protections & 17.2\% & 53.8\% & 57.1\% & \textbf{58.6\%} & 57.9\% & 58.0\% \\
 General Information  & 23.4\% & 85.7\% & 81.7\% & 82.0\% & \textbf{87.2\%} & 81.2\% \\
 Knowledge & 18.8\% & 75.6\% & \textbf{81.4\%} & 71.6\% & 80.9\% & 81.0\% \\
 Material Adverse Effect & 14.5\% & 44.0\% & 47.7\% & 49.8\% & 48.8\% & \textbf{50.9\%} \\
 Operating and Efforts Cov. & 22.0\% & 84.8\% & 85.7\% & \textbf{89.0\%} & 86.9\% & 86.6\% \\
 Remedies & 10.9\% & 88.2\% & 94.3\% & \textbf{100\%} & 96.6\% & 95.0\% \\
 \hline
 Overall & 16.8\% & 52.6\% & 55.5\% & 55.9\% & 57.1\% & \textbf{57.8\%} \\
 \bottomrule
 \end{tabular}}
 
 \caption{Single-task AUPR scores for each deal point category and fine-tuned model.
 Scores are calculated over the test split, which includes main and abridged examples but not rare answer examples.
 Each category score is calculated as the mean minority-class AUPR over all questions in the category and over three runs.
 The overall score is the mean AUPR score over all questions (not the mean over categories).
 See Appendix \ref{appendix:categories} for category descriptions.
 }
 \label{table:performance_single}
 \end{table*}

\begin{table*}[htbp]
\vspace{-7pt}
\setlength\tabcolsep{11pt}
\centering
\scalebox{0.85}{\begin{tabular}{l|cccccc}
\textbf{Deal Point Category} & \textbf{Random} & \textbf{RoBERTa} & \hspace{-0.1cm}\textbf{LegalBERT}\hspace{-0.1cm} & \textbf{DeBERTa} \\
\hline
Conditions to Closing & 20.4\% & 40.3\% & \textbf{46.2\%} & \textbf{46.2\%} \\
Deal Protections & 17.2\% & 48.6\% & \textbf{53.6\%} & 53.0\% \\
General Information  & 23.4\% & \textbf{80.2\%} & 74.8\% & 67.7\% \\
Knowledge & 18.8\% & 68.3\% & \textbf{73.0\%} & 71.8\% \\
Material Adverse Effect & 14.5\% & 48.3\% & \textbf{50.7\%} & 47.8\% \\
Operating and Efforts Cov. & 22.0\% & 80.3\% & \textbf{87.3\%} & 74.2\% \\
Remedies & 10.9\% & 51.0\% & \textbf{83.6\%} & 77.9\% \\
\hline
Overall & 16.8\% & 51.4\% & \textbf{55.8\%} & 53.0\% \\
\bottomrule
\end{tabular}}

\caption{Multi-task AUPR scores for each deal point category and fine-tuned model. We omit BERT because the architecturally similar RoBERTa and LegalBERT outperform BERT in the single-task setting, and omit BigBird due to compute limitations.
}
\label{table:performance_multi}
\end{table*}

\section{Experiments}

\subsection{Setup}

\paragraph{Metrics.} Many MAUD questions have an imbalanced answer distribution, so we use area under the precision-recall curve (AUPR) as our primary metric.
See Appendix \ref{appendix:aupr_details} for details on how we average AUPR across different questions and answers.

\paragraph{Models.} 
We fine-tune both single-task and multi-task pretrained language models on MAUD using the Transformers library \cite{wolf-etal-2020-transformers}.
In the single-task setting, we fine-tune individual pretrained LLMs for every deal point question.
We evaluate the single-task performance of BERT-base (110M params), RoBERTa-base (125M params), LegalBERT-base (110M params), DeBERTa-v3-base (184M params), and BigBird-base (127M params).

In the multi-task setting, we fine-tune LLMs with one classification head for every deal point question. We evaluate the multi-task performance of RoBERTa-base, LegalBERT-base, and DeBERTa-v3-base.
See Appendix \ref{appendix:training} for full training details.

BERT \citep{Devlin2019BERTPO} is a bidirectional Transformer that established state-of-the-art performance on many NLP tasks. LegalBERT \citep{chalkidis_legal_2020} pretrains BERT on a legal corpus. RoBERTa \citep{Liu2019RoBERTaAR} improves on BERT, using the same architecture, but pretraining on an order of magnitude more data. DeBERTa \citep{He2020DeBERTaDB} improves upon RoBERTa by using a disentangled attention mechanism and more parameters.

27.6\% of the unique deal point texts in MAUD and 50.0\% of texts across all examples are longer than 512 RoBERTa-base tokens, motivating our evaluation of BigBird-base.
BigBird \citep{zaheer2020big} is initialized with RoBERTa and trained on longer input sequences up to 4,096 tokens using a sparse attention pattern that scales linearly with the number of input tokens. No deal point texts in MAUD have more than 4,096 tokens. For all models except BigBird we truncate texts to the first 512 tokens.

\subsection{Results}

Our baseline models achieved high AUPR scores in the Remedies, General Information, and Operating \& Efforts Covenant categories, but scored lower on other categories, particularly
Deal Protections \& Related Provisions (best single-task AUPR 58.6\%),
Conditions to Closing (48.2\%),
and Material Adverse Effect (50.9\%).
Our results indicate that there is substantial room for improvement in these three hardest categories, which have the longest text lengths
(see Table \ref{table:dataset_statistics})
and which attorneys also find to be the most difficult to review.

Generally, larger and newer models had higher mean performance on MAUD. In the single-task setting, DeBERTa achieved an overall score of 57.1\% AUPR, compared with 55.5\% for RoBERTa and 52.6\% for BERT. BigBird achieved the highest score of 57.8\% AUPR, slightly outperforming DeBERTa.
See Tables \ref{table:performance_single} and \ref{table:performance_multi} for full results on single-task and multi-task models.

\begin{figure}
    \centering
    \includegraphics[width=0.48\textwidth]{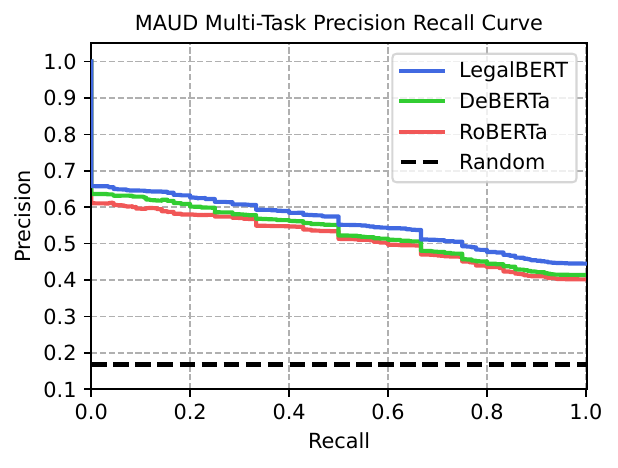}
    \vspace{-0.60cm}
    \caption{Precision-recall curves for multi-task models averaged over all MAUD questions.}
    \vspace{0.05cm}
    \label{fig:pr_curves}
\end{figure}

\begin{figure}
    \centering
    \vspace{0.1cm}
    \includegraphics[width=0.465\textwidth]{"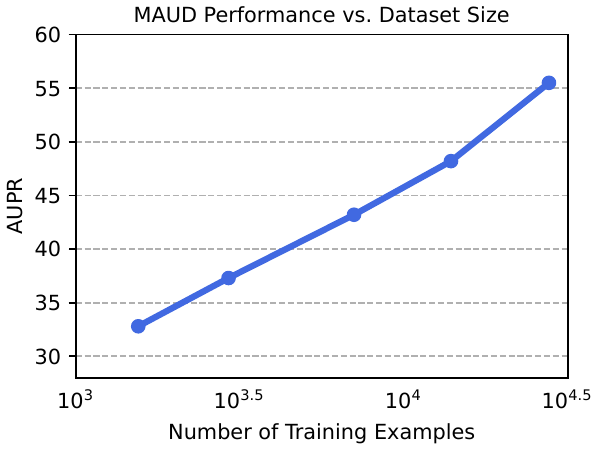"}
    \caption{RoBERTa-base AUPR as a function of the number of training examples, highlighting the value of our dataset's size. AUPR is averaged over three runs.}
    \vspace{-0.38cm}
    \label{fig:size_ablation}
\end{figure}

\paragraph{Single-Task versus Multi-Task Performance.}
RoBERTa and DeBERTa single-task models outperformed their multi-task counterparts by about 4 pp AUPR. However, for LegalBERT these models had approximately the same performance.

\paragraph{Effect of Pretraining on Legal Corpus.} In the single-task setting, LegalBERT outperforms BERT and slightly outperforms RoBERTa. In the multi-task setting, LegalBERT also outperforms DeBERTa. The strong performance of LegalBERT suggests that that pretraining on legal data is helpful for MAUD. 
See Figure \ref{fig:pr_curves} for a precision-recall curve comparing multi-task LegalBERT to other models.

\paragraph{F1 Scores}

Tables \ref{table:f1_micro} and \ref{table:f1_macro} present micro- and macro-averaged F1 scores for our multi-task models. LegalBERT has the highest micro F1 score in all categories excluding Knowledge, and the highest macro F1 score in all but three categories.

\paragraph{Dataset Size Ablation.}

We trained single-task RoBERTa models on random subsets of MAUD training data to evaluate the effect of dataset size on performance (see Figure \ref{fig:size_ablation}).
We found that RoBERTa models trained on all training examples had an overall AUPR score 7.3 percentage points higher than those trained on a 50\% subset of the dataset and 23.7 percentage points higher than models trained on only a 5\% subset.


\paragraph{Main vs. Abridged Datasets}
Table \ref{table:abridged_effect} reports multi-task AUPR scores over main and abridged examples separately.  As expected, the abridged subscores are higher than the main subscores.


\section{Conclusion}
MAUD is a large-scale expert-annotated dataset which facilitates NLP research on a specialized merger agreement review task, based on the American Bar Association's Public Target Deal Point Study.
MAUD can accelerate research towards specialized legal tasks like merger agreement review, while also serving as a benchmark for assessing NLP models in legal text understanding.
Fine-tuned LLM baselines exhibit strong performance on some deal point categories, but there is significant room for improvement on the three hardest categories.


\section{Ethics Statement}
\subsection{Data Collection} \label{section:demographics}
Our data was created by volunteer annotators from a non-profit legal organization, who joined the organization in order to create this dataset. None of our annotators were compensated monetarily for their time.
Among our 36 annotators, 20 were male and 16 were female. 33 annotators are based in the United States and 3 annotators are based in Europe.

\subsection{Societal Impact}
Advances in ML contract review, including merger agreement review, can reduce the costs of and increase the availability of legal services to businesses and individuals. In coming years, M\&A attorneys would likely benefit from having auxiliary analysis provided by ML models.



\subsection{Limitations}
MAUD enables research on models that can automate a specialized labelling task in the ABA Study, but does not target the other task performed in the ABA Study, which is the extraction of deal point texts from merger agreements.

We reserve this task for future work. For researchers interested in the deal point extraction task, we also release the 152 original contract texts and span annotations. Details on the span annotations and a preliminary baseline can be found in Appendix \ref{appendix:extraction}.

The 152 merger agreements in MAUD involve the acquisitions of most but not all of the U.S. public target companies exceeding \$200 million in value that were closed in 2021. 
Merger agreements for private companies or public companies that do not exceed \$200 million in value are not included, and consequently models trained on MAUD may be less performant for deal point texts extracted for these merger agreements.

The deal point questions and the list of predefined deal point answers to each question were created by experienced M\&A attorneys and standardized by the ABA, but they do not represent all of the deal points that are important in a merger agreement. MAUD should not be used as the sole source for developing AI tools for merger agreement review and drafting.

Many deal point texts exceed the maximum sequence lengths of our baseline models, and therefore we truncated texts to 512 tokens in all models except BigBird.

BigBird has a large GPU memory footprint (see Appendix \ref{appendix:training}).
Furthermore, prior work has shown that Longformer models (similar sparse attention mechanism) does not significantly outperform a 1024-token BART model on long texts, even though the BART model must truncate the long text~\cite{shaham2022scrolls}.
Future research could apply other methods for enabling longer sequence lengths like ALiBi~\cite{alibi} and FlashAttention~\cite{flash_attention} to improve performance on long deal point texts.

Our multi-task model adds a new classification head for every question. Using the extended deal point question descriptions (see Appendix \ref{appendix:nl_examples}), future work may be able to train multi-task models that encode the question description as an input and scale to an arbitrary number of questions, or query answers from generative models using the extended descriptions for prompting.

\newpage
%
\bibliography{main}
\bibliographystyle{acl2022_natbib}

\newpage
\appendix
\section{Appendix}

\begin{table*}[htbp]
\centering
\scalebox{0.65}{
\begin{tabular}{ccccc}
\hline
\textbf{contract\_name}          & \textbf{category} & \textbf{text} & \textbf{question} & \textbf{answer} \\
\hline
contract\_93 & Material Adverse Effect &
\makecell[l]{“Company Material Adverse Effect” \\ shall mean any state of facts, change, \\ condition, occurrence, effect, event, ...} 
& FLS (MAE) Standard-Answer & 

\makecell[l]{\CheckedBox “Would” (reasonably) be expected to \\ \Square “Could” (reasonably) be expected to \\ \Square Other forward-looking standard}
 \\
 \hline
contract\_102 & General Information & \makecell[l]{(i) each share of Company Common \\ Stock (including each share of \\ Company Common Stock described ...}
& Type of Consideration &
\makecell[l]{\CheckedBox All Cash \\ \Square All Stock \\ \Square Mixed Cash/Stock} \\
 \hline
contract\_77 & Conditions to Closing &
\makecell[l]{Section 3.1 Organization, Standing \\ and Power. \textless omitted\textgreater 
Section 3.2 \\ Capital Stock. \textless omitted\textgreater (b) All \\ outstanding shares of capital stock \\ and other voting securities or ...}

 & \makecell{Accuracy of \\ Fundamental Target \\ R\&Ws-Types of R\&Ws} &
  \makecell[l]{ \CheckedBox Capitalization-Other \\ \CheckedBox Authority \\ \CheckedBox Approval \\ \hspace{4mm} \vdots \\ \Square Other} 
 \\
\hline
\end{tabular}}

\caption{MAUD contains three CSV files corresponding to the train, dev, and test splits of the dataset. We illustrate some example rows in the table above, using a subset of the CSV columns. For full details on the dataset's format, we refer the reader to the MAUD Data Sheet or the dataset README.}
\label{table:dataset_structure}
\end{table*}
\subsection{Licensing}
MAUD is licensed under CC-BY 4.0.

\subsection{Original Merger Agreement Texts}
The 152 original merger agreement texts are available as text files in the supplementary data.

\subsection{Output Constraints} \label{appendix:output_constraints}
There are two conditions by which certain questions or answers in MAUD can be invalidated.

\paragraph{None-of-the-above Answer Constraint.} For any multi-label questions, if the “No” is label applied, then all other labels must be false. The five questions affected by this constraint are:

\begin{itemize}
    \item \texttt{W/N/A/F subject to "disproportionate impact"-Answer}
    \item \texttt{W/N/A/F applies to-Answer}
    \item \texttt{A/P/C application to-Answer}
    \item \texttt{FLS (MAE) Standard-Answer}
    \item \texttt{FLS (MAE) applies to}
\end{itemize}

\paragraph{Conditional Questions.}
Some questions are only applicable if another question had a particular answer. These three conditional questions follow:

\begin{itemize}
\item \texttt{Stock Deal: Fixed Ratio v. Fixed Value-Answer} is valid only if the answer to \texttt{Type of Consideration-Answer} is \texttt{"All Stock"} or \texttt{"Mixed Cash/Stock"}.
\item \texttt{Constructive Knowledge-Answer} which is valid only if \texttt{Knowledge Definition-Answer} is \texttt{"Constructive Knowledge"}.
\item \texttt{COR standard (board determination only)-answer} is valid only if \texttt{COR permitted with board fiduciary determination} is \texttt{"Yes"}.
\end{itemize}

If a conditional question is invalidated, then it does not appear in the dataset for the respective contract.

\begin{table*}[htbp]
\setlength\tabcolsep{11pt}
\centering
\scalebox{0.85}{\begin{tabular}{l|cccc}
\hline
\textbf{Data Subset} & \textbf{Random} & \textbf{RoBERTa} & \hspace{-0.1cm}\textbf{LegalBERT}\hspace{-0.1cm} & \textbf{DeBERTa} \\
\hline
Abridged & 22.1\% & 67.5\% & \textbf{73.3\%} & 64.3\%\\
Main & 23.1\% & 65.1\% & \textbf{70.0\%} & 63.5\% \\
\bottomrule
\end{tabular}}

\caption{Mean AUPR scores for multi-task models, computed separately for main test examples and for abridged test examples. For a fair comparison, we compute these scores only over questions that have any abridged examples.
As expected, abridged subscores are higher than main subscores.
}
\label{table:abridged_effect}
\end{table*}

%

\begin{table*}[htbp]
\setlength\tabcolsep{11pt}
\centering
\scalebox{0.85}{\begin{tabular}{l|cccccc}
\textbf{Deal Point Category} & \textbf{RoBERTa} & \hspace{-0.1cm}\textbf{LegalBERT}\hspace{-0.1cm} & \textbf{DeBERTa} \\
\hline
Conditions to Closing & 66.0\% & \textbf{66.5\%} & 64.0\% \\
Deal Protections & 65.5\% & \textbf{67.1\%} & 65.3\% \\
General Information  & 85.5\% & \textbf{86.0\%} & 82.8\% \\
Knowledge & \textbf{88.7\%} & 87.9\% & 87.9\% \\
Material Adverse Effect & 79.7\% & \textbf{81.0\%} & 76.6\% \\
Operating and Efforts Cov. & 87.5\% & \textbf{89.9\%} & 83.0\% \\
Remedies & 90.6\% & \textbf{97.4\%} & 94.3\% \\
\hline
Overall & 74.8\% & \textbf{76.1\%} & 72.8\% \\
\bottomrule
\end{tabular}}

\caption{Micro-averaged F1 scores for each deal point category and multi-task model. LegalBERT, the most
performant multi-task model by AUPR score (see Table \ref{table:performance_multi}), has the highest micro-averaged F1 scores in every category except Knowledge.
}
\label{table:f1_micro}
\end{table*}
\begin{table*}[htbp]
\setlength\tabcolsep{11pt}
\centering
\scalebox{0.85}{\begin{tabular}{l|cccccc}
\textbf{Deal Point Category} & \textbf{RoBERTa} & \hspace{-0.1cm}\textbf{LegalBERT}\hspace{-0.1cm} & \textbf{DeBERTa} \\
\hline
Conditions to Closing & 41.9\% & 41.9\% & \textbf{44.0\%} \\
Deal Protections & 50.9\% & \textbf{52.8\%} & 50.7\% \\
General Information  & \textbf{76.3\%} & 68.4\% & 61.1\% \\
Knowledge & \textbf{76.5\%} & 75.4\% & 75.3\% \\
Material Adverse Effect & 61.6\% & \textbf{62.3\%} & 60.1\% \\
Operating and Efforts Cov. & 79.1\% & \textbf{82.6\%} & 73.2\% \\
Remedies & 68.6\% & \textbf{91.7\%} & 84.7\% \\
\hline
Overall & 53.3\% & \textbf{59.7\%} & 57.3\% \\
\bottomrule
\end{tabular}}

\caption{Macro-averaged F1 scores for each deal point category and multi-task model. LegalBERT has the highest F1 score in all but three categories. }
\label{table:f1_macro}
\end{table*}

\begin{table*}[htbp]
\centering
\scalebox{0.8}{
\begin{tabular}{lccc}
\hline
\textbf{Deal Point Category} 
&
\begin{tabular}{@{}c@{}}
\textbf{Deal Point} \\ \textbf{Questions}\end{tabular}
& 
\begin{tabular}{@{}c@{}}\textbf{Percent} \\
\textbf{Long Texts} \end{tabular}
\\
\hline
Conditions to Closing & 9 & 43.9\%
 \\
Deal Protection and Related Provisions & 31 & 21.7\%
 \\
General Information & 1 & 5.6\%
 \\
Knowledge & 3 & 16.7\%
 \\
Material Adverse Effect & 39 & 99.0\%
 \\
Operating and Efforts Covenant & 8 & 2.1\%
 \\
Remedies & 1 & 0.0\%
 \\
 \hline
All Categories & 92 & 50.0\%
 \\
 \hline
\end{tabular}}

\caption{Number of deal point questions and long text proportions by category. ``Percent Long Texts" refers to the proportion of annotations with deal point texts longer than 512 tokens when using a \texttt{roberta-base} tokenizer. Conditions to Closing, Deal Protection and Related Provisions, and Material Adverse Effect have the largest proportion of long texts.}
\label{table:dataset_statistics}
\end{table*}

\subsection{Training details} \label{appendix:training}

\paragraph{Training.}
We fine-tune both single-task and multi-task models using the AdamW optimizer 
\cite{loshchilov2018decoupled_adamw}
with weight decay $0.01$. We oversample to give every answer equal proportion.
We trained our final models on the combined training and development splits, averaging test AUPR scores over three runs.
For all models except BigBird we truncate deal point texts to 512 tokens.


\paragraph{Models.}
The BERT, RoBERTA, LegalBERT, DeBERTa-v3, and BigBird pretrained language models that we use in our experiments are available on HuggingFace Hub as \texttt{bert-base-cased}, \texttt{roberta-base}, \texttt{nlpaueb/legal-bert-base-uncased}, \texttt{microsoft/deberta-v3-base}, and \texttt{google/bigbird-roberta-base}.

\paragraph{Multi-Task Models.}

For multitask experiments, we attach $144$ classification heads to each model, one for each question (including each multilabel binary question) in the dataset.

For every question we maintain a different shuffled queue of training examples. Each training batch fed to the classifier consists of $16$ training examples drawn in round-robin order from the question queues.

\paragraph{Hyperparameter Selection.}
For single-task BERT, RoBERTa, LegalBERT, and DeBERTa-v3 experiments, including the RoBERTa dataset size ablation experiment, we use batch size $16$. We grid-searched over learning rates $\{1 \times 10^{-5}, ~3 \times 10^{-5}, ~1 \times 10^{-4} \}$ and number of updates $\{100, 200, 300, 400\}$.

For single-task BigBird experiments we use batch size $8$. We grid-searched over learning rates $\{1 \times 10^{-5}, 1 \times 10^{-4} \}$ and number of updates $\{200, 400, 600, 800\}$.

For all multi-task models models, we used batch size $16$, and grid-searched over learning rates $\{1 \times 10^{-5}, ~3 \times 10^{-5}, ~1 \times 10^{-4} \}$ and number of epochs $\{1, 2, 3, 4, 5, 6\}$. Validation AUPR scores were averaged over 3 runs.

\paragraph{Infrastructure and Computational Costs.}
We trained BERT and RoBERTa experiments in parallel on A5000 GPUs, using about 12GB of GPU memory. Three runs of fine-tuning models for every question with 400 updates took about one GPU-day per learning rate setting.

We trained DeBERTa-v3 experiments in parallel on A4000 GPUs, using about 20GB of GPU memory. Three runs of fine-tuning models for every question with 400 updates took about two GPU-days per learning rate setting.

We trained BigBird experiments in parallel on A4000, A5000, and A100 GPUs, choosing the minimum GPU size required to accomodate the GPU usage of the single-task model, which varied with each model's maximimum deal point text length. The single-task models corresponding to questions with the longest deal point texts required about 75 GB of GPU memory. Three runs of fine-tuning models for every question with 800 updates took 4 to 5 GPU-days per learning rate setting.

Multi-task models were trained on an A100 GPU and used more memory than their single-task equivalents. Three runs of 6 epochs of training took about 6 GPU-hours per learning rate setting and model.

\subsection{Details on MAUD AUPR Score} \label{appendix:aupr_details}
For every question, we calculate the minority-class AUPR score for each answer and then average to get a mean AUPR score for the question. Then we average over all question scores to get an overall AUPR score for a model.


For example, consider a
deal point question $Q$, with three possible answers: $A1$, $A2$, and $A3$, which have $50$, $10$, and $10$ test examples respectively. For the unique question-answer pair $(Q, A1)$, we first binarize all answers as $A1$ or $\neg A1$. The minority binarized answer is $\neg A1$, with 20 examples, and so the AUPR score for $(Q, A1)$ is calculated using positive class $\neg A1$. To get the AUPR score for question $Q$, we average the AUPR scores for $(Q, A1)$, $(Q, A2)$, and $(Q, A3)$.



\subsection{Best-Performing Hyperparameters}
For brevity we present the over 300 combinations of best hyperparameters as CSV files in the supplementary materials.

\subsection{Evaluation Variability}
We find that the average overall AUPR over three runs for our models can vary by 1-2\%.

\subsection{Extended Descriptions for Deal Point Questions and Types}\label{appendix:nl_examples}
The supplementary materials include extended descriptions of each deal point type. These
extended descriptions may be useful for prompting and providing additional context 
 to models trained on MAUD. For the Material Adverse Effect category we also include extended descriptions for each
 deal point question and release these descriptions along with the dataset.

Table \ref{table:nl_example_dp_questions} and Table \ref{table:nl_example_dp_types} present example descriptions of deal point questions and types.

\begin{table*}[htbp]
    \centering
    \scalebox{0.85}{
    \begin{tabular}{|p{0.4\textwidth}|p{0.5\textwidth}|}
    \hline
    \textbf{Deal Point Question} & \textbf{Description} \\ \hline
    Accuracy of Target "General" R\&W: Bringdown Timing Answer & When does the general representations and warranties need to be accurate? Does the general representations and warranties need to be accurate at closing or both at signing and at closing? \\ \hline
    Type of Consideration-Answer & Is the type of consideration all cash, all stock, mixed cash/stock or mixed cash/stock: election? \\ \hline
    Stock Deal: Fixed Ratio v. Fixed Value-Answer & Is the consideration based on fixed ratio or fixed value? \\ \hline
    MAE definition includes reference to Target "prospects" (Y/N) & Does the MAE definition includes reference to the word "prospect" or "prospects"? \\ \hline
    "Ability to consummate" concept is subject to MAE carveouts & Is the impact on the ability to consummate subject to MAE carveouts? \\ \hline
    \end{tabular}}
    \caption{Example Deal Point Question descriptions. We are in the process of writing prompt-friendly descriptions for every deal point question. These descriptions will be released with the MAUD dataset.}
    \label{table:nl_example_dp_questions}
\end{table*}

\begin{table*}[htbp]
    \centering
    \scalebox{0.85}{
    \begin{tabular}{|p{0.4\textwidth}|p{0.5\textwidth}|}
    \hline
    \textbf{Deal Point Type} & \textbf{Description} \\ \hline
    Ordinary course covenant & Whether the acquisition agreement requires the target company to continue its ordinary course of business, how is "ordinary course" defined and the exceptions to this obligation? \\ \hline
    Negative interim operating covenant & Whether the acquisition agreement prohibits the target company from taking certain actions and the exceptions to this obligation? \\ \hline
    General Antitrust Efforts Standard & The level of efforts required to obtain anti-trust clearance for the acquisition \\ \hline
    Limitations on Antitrust Efforts & The limitations to the efforts required to obtain anti-trust clearance for the acquisition \\ \hline
    Specific Performance & Whether in the event of a breach of the acquisition agreement, the non-breaching party is automatically entitled to specific performance? \\ \hline
    \end{tabular}}
    \caption{Example Deal Point Type descriptions. All 22 Deal Point Type descriptions can be found in the supplementary materials and will be released
    with the MAUD dataset.}
    \label{table:nl_example_dp_types}
\end{table*}

\subsection{Example Annotations in the Datasets}\label{appendix:example}

Table \ref{table:dataset_structure} shows the dataset structure as well as a few example annotations.

%
%

\subsection{Other Dataset Statistics}

Table \ref{table:dataset_statistics} shows the percentage of deal point texts that are longer than 512 tokens and the number of deal point questions in each category.

\subsection{Category Descriptions}\label{appendix:categories}

We describe the seven categories of deal points found in our dataset.
\begin{enumerate}
    \item{\textit{General Information}. This category includes the type of consideration and the deal structure of an acquisition.}
    \item{\textit{Conditions to Closing}. This category specifies the conditions upon the satisfaction of which a party is obligated to close the acquisition. These conditions include the accuracy of a target company’s representations and warranties, compliance with a target company’s covenants, absence of certain litigation, absence of exercise of appraisal or dissenters rights, absence of material adverse effect on the target company.}
    \item{\textit{Material Adverse Effect}. This category includes a number of questions based on the Material Adverse Effect definition. Material Adverse Effect defines what types of event constitutes a material adverse effect on the target company that would allow the buyer to, among other things, terminate the agreement.}
    \item{\textit{Knowledge}. This category includes several questions based on the definition of Knowledge. Knowledge defines the standard and scope of knowledge of the individuals making representations on behalf of the target companies.}
    \item{\textit{Deal Protection and Related Provisions}. This category describes the circumstances where a target company’s board is permitted to change its recommendation or terminate the merger agreement in order to fulfill its fiduciary obligations.}
    \item{\textit{Operating and Efforts Covenants}. This category includes requirements for a party to take or not to take specified actions between the signing of the merger agreement and closing of the acquisition. The types of covenants include obligation to conduct business in the ordinary course of business and to use reasonable efforts to secure antitrust approval.}
    \item{\textit{Remedies}. This category describes whether a party has the right to specific performance.}
\end{enumerate}

\subsection{Labeling Process}\label{appendix:labeling process}
MAUD is a collective effort of over 10,000 hours by law students, experienced lawyers, and machine learning researchers.
Prior to labeling, each law student attended 70-100 hours of training that included live and recorded lectures by experienced M\&A lawyers and passing multiple quizzes.
Law students also read an instructions handbook.

Our volunteer annotators are experienced lawyers and law students who are part of a non-profit legal organization. None of the volunteers were compensated monetarily for their time.
See Section \ref{section:demographics} for annotator demographics.

\paragraph{Data Verification.}
The law students who annotated MAUD worked in teams of three. Each annotation in the main and abridged datasets was first annotated collectively, by a consensus established by a law student team. This annotation includes both the text extraction and the deal point answers. When the team of law students could not reach an agreement on an annotation, they escalated to an experienced M\&A lawyer. Finally, every law student annotation was reviewed by an experienced M\&A lawyer for accuracy.

We unfortunately did not retain the records necessary for us to calculate inter-annotator agreement metrics. However, lawyers reviewing the student annotations report that they agreed about 80\% of the time with student annotations.

\subsection{Data Quality}
We conducted a post-hoc quality check of deal point answers in one of the most difficult categories, Material Adverse Effect (MAE), with the help of an M\&A attorney who was not involved in the original annotation process.

The attorney answered the MAE questions for $10$ randomly selected contracts and disagreed with our gold labels $3$ times out of a total of $440$ labels. After conferring with our annotation team about these disagreements, the attorney decided that all three answers would be better answered by the gold label.

\paragraph{Main and Abridged Datasets.}
To create the main dataset and the abridged dataset, the law students conducted manual review and labeling of the merger agreements uploaded in eBrevia, an electronic contract review tool. On a periodic basis, the law students exported the annotations into reports, and sent them to experienced lawyers for a quality check. The lawyers reviewed the reports or the labeled contracts in eBrevia, provided comments, and addressed student questions.

Where needed, reviewing lawyers escalated questions to a panel of 3-5 expert lawyers for discussions and reached consensus. Students or the lawyers made changes in eBrevia accordingly.

\paragraph{Rare Answers Dataset.}
To create the rare answers dataset, legal experts copied an example from the main dataset and minimally edited the deal point text to create an example with a rare answer. These edits were then reviewed by an experienced attorney to ensure accuracy.


For example, the deal point question ``Limitations on Antitrust Efforts" originally had very few examples of ``Dollar-based standard" deal point answer. To create examples with this rare answer, the annotators changed phrases in the deal point text similar to ``no obligation to divest or take other actions" with language implying a dollar-based standard, such as ``Remedy Action or Remedy Actions with assets which generated in the aggregate an amount of revenues that is in excess of USD 50,000,000."


\paragraph{Final Annotation Formatting.}
We exported the final annotations as three CSV files corresponding to the main, abridged, and rare answers datasets. For example rows in the dataset, see Table \ref{table:dataset_structure}.

\subsection{Extraction Dataset and Baseline}
\label{appendix:extraction}
We also release a deal point extraction dataset in SQuAD2.0 JSON format~\cite{rajpurkar-etal-2018-know}, as commonly used for Extractive Question Answering datasets. The extraction dataset and the training and evaluation code for a \texttt{roberta-base} baseline
\iftoggle{release}{
    can be found at \href{http://github.com/TheAtticusProject/maud-extraction}{github.com/TheAtticusProject/maud-extraction}.
}{
    can be found in the supplementary materials and will be released on GitHub.
}

The annotated contract spans in this extraction dataset correspond to deal point texts in the primary reading comprehension task's main dataset.

The deal point texts extracted by the annotators are often discontiguous, with gaps ranging from a sentence to several pages in length.

\paragraph{Deal Point Types.}
As described in Section \ref{section:dataset}, each example in MAUD's primary reading comprehension task contains an extracted deal point text, a deal point question that can be asked of this text, and a deal point category.

Another field in each example is the deal point type. There are 22 deal point types, and each deal point type belongs to exactly one deal point category. The same set of questions are asked of all deal points with the same type.

In MAUD's extraction task, the Extractive QA model is asked to identify spans of text inside the full contract text that were annotated by legal experts as belonging to a particular deal point type.

See Table \ref{table:extraction_types} for a list of deal point types and the number of examples of each type.


\paragraph{Extraction Data Formatting.}
For every deal point type and every contract, we format the question as follows:
``Highlight the parts of the text (if any) related to ``\textless Deal Point Type\textgreater" that should be reviewed by a lawyer."

\paragraph{Extraction Dataset Splits.}
We build train, dev, and test splits by splitting the 152 contracts in a 80-10-10 ratio. The deal point texts as extracted by the annotators are often discontiguous, with gaps ranging from a sentence to several pages in length. Therefore the number of spans of each type can be exceed than the number of questions.


\begin{table*}[htbp]
        \centering
        \scalebox{0.8}{
        \begin{tabular}{l|ccc|c}
        \hline
        \textbf{Deal Point Type} &
        \textbf{Train}&
        \textbf{Dev} &
        \textbf{Test} &
        \textbf{RoBERTa-base (AUPR)} \\
        \hline
        Absence of Litigation Closing Condition & 25 & 6 & 2 & 5.8\% \\
        Accuracy of Target R\&W Closing Condition & 1,120 & 153 & 161 & 29.6\% \\
        Agreement provides for matching rights in connection with COR & 469 & 67 & 53 & 14.6\% \\
        Agreement provides for matching rights in connection with FTR & 424 & 56 & 46 & 14.8\% \\
        Breach of Meeting Covenant & 102 & 14 & 6 & 4.3\% \\
        Breach of No Shop & 269 & 28 & 35 & 18.6\% \\
        Compliance with Covenant Closing Condition & 231 & 32 & 26 & 86.7\% \\
        FTR Triggers & 241 & 45 & 27 & 44.0\% \\
        Fiduciary exception to COR convent & 454 & 58 & 50 & 31.4\% \\
        Fiduciary exception: Board determination (no-shop) & 276 & 43 & 35 & 47.9\% \\
        General Antitrust Efforts Standard & 181 & 27 & 22 & 33.6\% \\
        Intervening Event Definition & 114 & 15 & 12 & 73.9\% \\
        Knowledge Definition & 121 & 16 & 16 & 100.0\% \\
        Limitations on FTR Exercise & 385 & 47 & 45 & 58.8\% \\
        MAE Definition & 367 & 59 & 40 & 3.2\% \\
        Negative interim operating convenant & 169 & 25 & 22 & 61.2\% \\
        No-Shop & 302 & 45 & 39 & 8.9\% \\
        Ordinary course covenant & 167 & 28 & 26 & 63.2\% \\
        Specific Performance & 134 & 18 & 20 & 81.5\% \\
        Superior Offer Definition & 183 & 30 & 23 & 61.4\% \\
        Tail Period \& Acquisition Proposal Details & 360 & 47 & 40 & 16.7\% \\
        Type of Consideration & 217 & 26 & 21 & 69.9\% \\
        \hline
        Overall & 6,311 & 885 & 767 & 19.7\%\\
        \hline
    \end{tabular}}

\caption{Span counts and RoBERTa-base AUPR for the different splits of the MAUD extraction dataset, grouped by deal point type. The overall deal point scores are calculated using the PR curve over all spans (not average over AUPR scores).
}
\label{table:extraction_types}
\end{table*}

\paragraph{Metrics.}
MAUD's extraction task is very similar to the contract review extraction task from \citet{hendrycks_cuad_2021}. In both tasks there is a large imbalance between the number of negative examples and positive examples in each contract. Consequently, we use the area under the precision recall curve (AUPR), averaged over three runs, as our primary metric.
Following \citet{hendrycks_cuad_2021}, we consider a candidate span from our model to be a match for a lawyer-annotated span if the Jaccard similarity index is at least $50\%$.

\paragraph{Training Setup.}
Since the most spans in the contract text are negative examples, we oversample positive examples to create a balanced training dataset.

We fine-tune a \texttt{roberta-base} model on the combined train and dev datasets using an A100 GPU.
We use Adam optimizer~\cite{adam} and batch size $40$. We use validation AUPR to select the best learning rate from $\{ 1 \times 10^{-5}, 3 \times 10^{-5}, 1 \times 10^{-4} \}$ and the best number of training epochs from $\{ 4, 6, 8\}$.
The best learning rate was $1 \times 10^{-4}$ and the best number of epochs was $4$.
We average our test AUPR score over three runs.

\paragraph{Results.}
Our RoBERTa model has an AUPR score of $19.7\%$.
This is far lower than the $42.6\%$ baseline AUPR score achieved by RoBERTa in \citet{hendrycks_cuad_2021}, suggesting that our contract review extraction task is much more challenging.

\paragraph{Limitations.}
Our RoBERTa model can only process windows of up to 512 tokens in size. However, for some deal point types, including Material Adverse Effect, most deal point texts are longer than 512 tokens. Our baseline model does not aggregate adjacent span predictions. If a gold-label span is more than twice as long as RoBERTa's context window, then it's impossible to for the model to predict a matching span with at least 50\% Jaccard similarity. Future work can explore models which longer context windows or which can aggregate span extraction across multiple context windows to improve performance on these deal point spans.

\end{document}